\documentclass{article} % For LaTeX2e
\usepackage{iclr2024_conference,times}

\usepackage{amsmath,amsfonts,bm}

\def\eqref#1{equation~\ref{#1}}
\def\1{\bm{1}}

\DeclareMathAlphabet{\mathsfit}{\encodingdefault}{\sfdefault}{m}{sl}
\SetMathAlphabet{\mathsfit}{bold}{\encodingdefault}{\sfdefault}{bx}{n}

\usepackage{hyperref}
\usepackage{url}
\usepackage{booktabs}
\usepackage{diagbox}
\usepackage{multirow}
\usepackage{array}
\usepackage{arydshln}
\usepackage{graphicx}
\usepackage{tabularx}
\usepackage{xcolor}
\usepackage{footmisc}

\title{Improving Discriminative Multi-Modal Learning with Large-Scale Pre-Trained  Models}

\author{Chenzhuang Du\textsuperscript{1}, Yue Zhao\textsuperscript{2}, Chonghua Liao\textsuperscript{1}, Jiacheng You\textsuperscript{1}, Jie Fu\textsuperscript{3}, Hang Zhao\textsuperscript{1 4 \dag} \\
IIIS, Tsinghua University\textsuperscript{1},
Peking University\textsuperscript{2},
HKUST\textsuperscript{3},
Shanghai Qi Zhi Institute\textsuperscript{4}\\
\texttt{ducz20@mails.tsinghua.edu.cn} \\
}

\iclrfinalcopy % Uncomment for camera-ready version, but NOT for submission.
\begin{document}

% \footnotetext[ ]{\dag: advisor of the project}
{\renewcommand{\thefootnote}{} \footnotetext{\dag: advisor of the project.}}

\maketitle
\begin{abstract}
% The abstract paragraph should be indented 1/2~inch (3~picas) on both left and
% right-hand margins. Use 10~point type, with a vertical spacing of 11~points.
% The word \textsc{Abstract} must be centered, in small caps, and in point size 12. Two
% line spaces precede the abstract. The abstract must be limited to one
% paragraph.

This paper investigates how to better leverage large-scale pre-trained uni-modal models to further enhance discriminative multi-modal learning.
% When applying these models to relatively smaller multi-modal datasets, we observe: 
% (1) 
% Using these pre-trained uni-modal models for multi-modal learning significantly improves performance. 
% Even when fine-tuned the models with only uni-modal data, they can outperform previous multi-modal models on some tasks.
Even when fine-tuned with only uni-modal data, these models can outperform previous multi-modal models in certain tasks. 
It's clear that their incorporation into multi-modal learning would significantly improve performance.
However, multi-modal learning with these models still suffers from insufficient learning of uni-modal features, which weakens the resulting multi-modal model's generalization ability.
While fine-tuning uni-modal models separately and then aggregating their predictions is straightforward, it doesn't allow for adequate adaptation between modalities, also leading to sub-optimal results.
% Training uni-modal models separately and then combining their predictions~(Uni-Modal Ensemble) is indeed a straightforward and effective method. However, without adaptation between the two modalities, this approach yields sub-optimal results.
To this end, we introduce \textbf{M}ulti-\textbf{M}odal \textbf{Lo}w-\textbf{R}ank \textbf{A}daptation learning (\textbf{MMLoRA}). 
% By freezing the weights of uni-modal fine-tuned models and adding extra trainable rank decomposition matrices to them, our method enhances adaptation between modalities and boosts overall performance.
By freezing the weights of uni-modal fine-tuned models, adding extra trainable rank decomposition matrices to them, and subsequently performing multi-modal joint training, our method enhances adaptation between modalities and boosts overall performance.
% This method freezes the weights of uni-modal fine-tuned models and adds trainable rank decomposition matrices at each layer to enable better adaptation between modalities, thereby enhancing performance.
% To this end, we propose Low-Rank Adaptation of Multi-Modal learning~(\textbf{MMLoRA}), which freezes the uni-modal finetuned model weights and injects trainable rank decomposition matrices to each layer.
% These trainable parameters assist the two modalities in adapting to each other in multi-modal training and improve overall performance.
We demonstrate the effectiveness of MMLoRA on three dataset categories: audio-visual (e.g., AVE, Kinetics-Sound, CREMA-D), vision-language (e.g., MM-IMDB, UPMC Food101), and RGB-Optical Flow (UCF101).

\end{abstract}
\section{Introduction}
% \textbf{Large-Scale Pre-trained Models are efficitive.}

Large-scale pre-trained models have exhibited remarkable performance across various downstream tasks~\citep{radford2021learning,brown2020language}. This trend has also been validated across different modalities, including language~\citep{openai2023gpt4,anil2023palm,touvron2023llama}, vision~\citep{radford2021learning,oquab2023dinov2}, and audio~\citep{girdhar2023imagebind,huang2022masked}. 
Building on this success, Multi-modal Large Language Models~(MLLM) have emerged. By connecting pre-trained language and vision models, MLLMs equip language models with the ability to `see'~\citep{dai2023instructblip,liu2023visual}. 
While their primary focus lies in text generation and dialogue, we delve into the utilization  of large-scale pre-trained models in discriminative multi-modal learning.
% The open-sourcing of some of these models has further fueled this momentum~\citep{cherti2023reproducible,girdhar2023imagebind,he2020deberta}, motivating us to delve deeper into exploring enhanced utilization of large-scale pre-trained models in discriminative multi-modal learning.

% \textbf{Applying those pre-trained models on discriminative multi-modal learning.}
Large-scale pre-trained models significantly enhance discriminative multi-modal learning performance~(as shown in Sec~\ref{sec_4_effect_largemodels}). In fact, simply fine-tuning these models with corresponding uni-modal data often outperforms recent multi-modal models. As Table~\ref{tab:ft-unimodal} illustrates, fine-tuning pre-trained vision and audio models enables uni-modal models to surpass the performance of previously proposed multi-modal models~\citep{fan2023pmr, peng2022balanced} on AVE, Kinetics-Sound, and CREMA-D. Similarly, for MM-IMDB, by fine-tuning an advanced language model, the uni-modal approach can surpass the performance of recent multi-modal models~\citep{li2023efficient}.

However, despite the power of large-scale pre-trained models, when applied to multi-modal joint training, they can lead to insufficient learning of uni-modal features. Specifically, during linear evaluation, encoders from multi-modal learning underperform compared to their uni-modal fine-tuned counterparts (as Table~\ref{tab:laziness} shows).
% However, despite the power of large-scale pre-trained models, their use in multi-modal joint training can also lead to insufficient learning of uni-modal features~(as Table~\ref{tab:laziness} shows). 
This issue is called Modality Laziness~\citep{du2023uni} or Modality Competition~\citep{huang2022modality}, which has been proven to affect the generalization performance of the resulting multi-modal models. Uni-Modal Ensemble (UME) provides a straightforward solution by simply aggregating predictions from separately learned uni-modal models.
% A straightforward approach, which is referred to as Uni-Modal Ensemble~(UME)~\citep{du2023uni}, is to fine-tune the individual pre-trained models using uni-modal data. Subsequent combination of predictions from different modalities yields the final multi-modal prediction. 
However, this approach lacks cross-modal interaction, which may result in inadequate learning of paired features~\citep{du2023uni} and subsequently lead to sub-optimal performance.

Considering that uni-modal fine-tuned models already capture a significant amount of features, we hypothesize only a limited number of parameters are required for cross-modal adaptation. To this end and taking inspiration from Parameter-Efficient Fine-Tuning, particularly LoRA~\citep{hu2021lora}, we introduce a novel approach called \textbf{M}ulti-\textbf{M}odal \textbf{Lo}w-\textbf{R}ank \textbf{A}daptation learning (\textbf{MMLoRA}). This method begins by freezing the weights of the uni-modal fine-tuned models and then introduces additional trainable rank decomposition matrices to a specific modality or all modalities' models.
Subsequently, it proceeds with multi-modal joint training. 
During joint training, these newly introduced parameters facilitate improved adaptation between modalities, resulting in collaborative enhancements in predictions. MMLoRA not only surpasses other methods but also outperforms its fully fine-tuned counterpart. We demonstrate the effectiveness of MMLoRA across three categories of multi-modal datasets: audio-visual datasets [AVE~\citep{tian2018audio}, Kinetics-Sound~\citep{arandjelovic2017look}, and CREMA-D~\citep{cao2014crema}], vision-language datasets [MM-IMDB~\citep{arevalo2017gated} and UPMC Food101~\citep{wang2015recipe}], and the RGB-Optical Flow action recognition dataset [UCF101~\citep{soomro2012ucf101}].

% \textbf{Our Methods}

% \textbf{Contributions.}

\begin{table}[t]
    \centering
    \caption{\textbf{Fine-tuning large-scale pre-trained models with uni-modal data.} We report the performance of fine-tuned uni-modal models and prior multi-modal models. The performance of the multi-modal models on AVE and CREMA-D is sourced from \citet{fan2023pmr}~(CVPR'23), on MM-IMDB from \citet{li2023efficient}~(CVPR'23) and on Kinetics-Sound from \citet{peng2022balanced}~(CVPR'22). The reported evaluation metrics are Top-1 Accuracy (AVE, Kinetics-Sound and CREMA-D) and F1-Micro/F1-Macro (MM-IMDB).
    We have \textbf{bolded} the performances of uni-modal models that outperform previous multi-modal models.}
    \vspace{3pt}
\begin{tabular}{c|c|c|c|c}
\toprule
   \diagbox{\textbf{Model}}{\textbf{Dataset}}  & \textbf{AVE} & \textbf{Kinetics-Sound} & \textbf{CREMA-D} & \textbf{MM-IMDB}  \\
\midrule
 Recent Multi-modal Model    & 68.1 & 63.1& 65.3& 66.7 / 61.7 \\
\midrule
 Fine-tuned Visual Model   & \textbf{88.1} & \textbf{84.3} & \textbf{77.7} & 60.2 / 52.6 \\
 Fine-tuned Audio / Language Model   & \textbf{85.6} & \textbf{69.6} &\textbf{75.8} & \textbf{68.6 / 63.9} \\
\bottomrule
\end{tabular}
    \label{tab:ft-unimodal}
\end{table}

\section{Related Work}

\textbf{Large-Scale Pre-Trained Models.} 
Models pre-trained on large-scale datasets have consistently demonstrated exceptional performance when fine-tuned for downstream tasks~\citep{devlin2018bert, yang2019xlnet, he2020deberta}. They have even showcased remarkable results in few-shot or zero-shot testing scenarios~\citep{openai2023gpt4, brown2020language, radford2021learning, anil2023palm}. Widely-used and effective pre-training methods include Generative Pre-Training~\citep{radford2018improving, chen2020generative}, Mask Data Modeling~\citep{he2022masked, devlin2018bert}, Contrastive Learning~\citep{radford2021learning, cherti2023reproducible, girdhar2023imagebind}, and others~\citep{gidaris2018unsupervised}. The release of certain model weights, like CLIP~\citep{cherti2023reproducible, radford2021learning}, ImageBind~\citep{girdhar2023imagebind}, and DeBERTa~\citep{he2020deberta}, has inspired us to employ them in our area of interest, discriminative multi-modal learning. Indeed, their integration has significantly enhanced the performance of various methods across different datasets.

\textbf{Discriminative Multi-Modal Learning.}
Multi-modal learning has been proven to be superior to uni-modal learning in various areas~\citep{zhang2023universal, xiao2020multimodal, huang2021makes}.
However, challenges like Modality Competition~\citep{huang2022modality} or Modality Laziness~\citep{du2023uni} often hinder the effectiveness of multi-modal joint training in sufficiently capturing uni-modal features~\citep{peng2022balanced, fan2023pmr, wu2022characterizing}. In practice, this has led to instances where multi-modal models empirically perform worse~\citep{wang2020makes}, even when uni-modal models employ the same encoder as the multi-modal counterparts.  
% As Table~\ref{tab:laziness} shows, when employing large-scale pre-trained models as encoders for individual modalities and subsequently engaging in multi-modal joint learning, similar issues  also exist. 
% While adding extra loss terms~\citep{fan2023pmr} or controlling the learning process through gradients~\citep{peng2022balanced} can alleviate this issue, the encoders in multi-modal joint learning always worse than their counterparts from uni-modal training.
% In response, \citet{du2023uni} introduces Uni-Modal Ensemble~(UME), which directly averages predictions of uni-modal models to derive the final prediction. While UME mitigates Modality Laziness, its performance is still sub-optimal due to the lack of cross-modal adaptation.
While some methods try to address this issue with extra loss terms~\citep{fan2023pmr} or gradient control~\citep{peng2022balanced}, encoders from multi-modal learning still underperform compared to uni-modal training. \citet{du2023uni} introduces Uni-Modal Ensemble (UME) to average uni-modal predictions, but it lacks cross-modal adaptation, limiting its performance.

\textbf{Parameter-Efficient Fine-Tuning (PEFT).}
Parameter-Efficient Fine-Tuning is a methodology designed to fine-tune only a subset of parameters in large language models, enhancing their adaptability to downstream tasks~\citep{houlsby2019parameter, hu2021lora, zhang2023adaptive, liu2023gpt, liu2021p, dettmers2023qlora}. Among these methodologies,  LoRA~\citep{hu2021lora} is the most notable, which introduces novel trainable rank decomposition matrices to facilitate adaptation to new tasks. Once training is completed, these parameters can be seamlessly merged with the existing ones without incurring any additional inference cost. 
Previous works have also adopted similar approaches, using a linear projection layer~\citep{liu2023visual} or a querying transformer~\citep{li2023blip,dai2023instructblip} to link a large visual model with a  large language model. However, these studies primarily focuses on textual dialogues or generation, while our paper mainly focus on  discriminative multi-modal learning.
% Drawing inspiration from LoRA, we introduce Multi-Modal Low-Rank Adaptation learning (MMLoRA). MMLoRA incorporates LoRA-like trainable parameters into uni-modal fine-tuned models, followed by multi-modal joint training. Through this collaborative training approach, these parameters facilitate cross-modality adaptation, resulting in enhanced overall performance.

\begin{table}[t]
    \centering
    \caption{\textbf{Selection of Pre-trained Encoder}. (1) All visual encoders are from OpenCLIP\citep{cherti2023reproducible}, where ViT-B is pre-trained on LAION-2B, and ViT-L is pre-trained on DataComp-1B. (2) All audio encoders are from Imagebind\citep{girdhar2023imagebind}. (3) The text encoders BERT and DeBERTa come from \citet{devlin2018bert} and \citet{he2020deberta} respectively. (4) The encoder for optical flow is a ResNet-18~\citep{he2016deep} pre-trained on ImageNet.}
    \vspace{3pt}
    \begin{tabular}{c|c|c|c|c|c|c}
    \toprule
       \diagbox{\textbf{Encoder}}{\textbf{Dataset}}  & \textbf{AVE}& \textbf{KS} & \textbf{CREMA-D} & \textbf{Food101} & \textbf{MM-IMDB} & \textbf{UCF101}  \\
    \midrule
     Visual    & ViT-B & ViT-B& ViT-B& ViT-L & ViT-L & ViT-B\\
     \midrule
      Audio/Text/Flow & ViT-B   & ViT-B & ViT-B & BERT-B & DeBERTa-L & ResNet-18 \\
      
    \bottomrule
    \end{tabular}
    \label{tab:encoder_select}
\end{table}

\begin{table}[t]
    \centering
    \caption{\textbf{Top-1 test accuracy (in \%) of linear evaluation on encoders from multi-modal training} and uni-modal
fine-tuned models on AVE, Kinetics-Sound and CREMA-D. The one with better performance is highlighted in \textbf{bold}.}
    \vspace{3pt}

    \begin{tabular}{c c c c c c c}
		\toprule
		\multirow{2}*{\textbf{Encoder Source}} & \multicolumn{2}{c}{\textbf{AVE}} & \multicolumn{2}{c}{\textbf{Kinetics-Sound}}& \multicolumn{2}{c}{\textbf{CREMA-D}}\\
		\cline{2-7} & RGB & Audio & RGB & Audio& RGB & Audio\\
		\midrule
		  Multi-Modal Training & 83.3 &  79.8 & 78.7 & 67.9 & 70.3 & 64.8 \\
	    % \cdashline{1-7}
            Uni-Modal Fine-Tuned & \textbf{88.1}  & \textbf{85.6} & \textbf{84.3} & \textbf{69.6} & \textbf{77.7} & \textbf{75.8}\\		
		\bottomrule
\end{tabular}
    \label{tab:laziness}
\end{table}

\section{Analysis and Method}

In this section, we first illustrate that even when employing large-scale pre-trained models for multi-modal joint training, they can still suffer from insufficient learning of uni-modal features.
We then introduce our proposed method, \textbf{M}ulti-\textbf{M}odal \textbf{Lo}w-\textbf{R}ank \textbf{A}daptation learning~(\textbf{MMLoRA}), to address this issue. 
\subsection{Multi-modal Learning with Large-Scale Pre-Trained Models}
\label{sec3_1_laziness}
Large-scale pre-trained models have demonstrated impressive performance in downstream tasks~\citep{radford2021learning}. 
With an increasing number of these models being open-sourced~\citep{cherti2023reproducible,girdhar2023imagebind,he2020deberta}, it motivates us to investigate how to further enhance discriminative multi-modal learning with them.

% \textbf{Encoder Selection.}
% For the datasets AVE, Kinetics-Sound, and CREMA-D, we use the ViT-B/16 from OpenCLIP~\citep{cherti2023reproducible} pre-trained on LAION-2B~\citep{schuhmann2022laion} as the Visual Encoder and the Audio Encoder from ImageBind~\citep{girdhar2023imagebind} as the Audio Encoder; For MM-IMDB, we use the ViT-L/14 from OpenCLIP pre-trained on DataComp-1B~\citep{gadre2023datacomp} as the Visual Encoder, and DeBERTa-Large~\citep{he2020deberta} as the Language Encoder. Note that these open-sourced models are merely encoders. 
Table~\ref{tab:encoder_select} displays the encoders chosen for various datasets. For a more comprehensive discussion and additional details, please refer to Sec~\ref{sec4_encoder_select}. It's important to note that for specific downstream tasks, an additional linear layer is necessary to map the extracted features to the label space.

\textbf{Finetuned pre-trained models with uni-modal data outperform previous multi-modal models.} 
We first directly fully fine-tune these pre-trained models using the uni-modal data, with results presented in Table~\ref{tab:ft-unimodal}. In these four datasets~(AVE, Kinetics-Sound, CREMA-D and MM-IMDB), fine-tuning a uni-modal model already outperforms the multi-modal models from recent publications~\citep{fan2023pmr,peng2022balanced,li2023efficient}. This encourages us to further investigate how to better utilize these models in discriminative multi-modal learning. 
% Note that the output of all pre-trained models is a feature; we need to add a linear layer to map these features to the prediction space.

\begin{figure}
    \centering
    \includegraphics[width=1\linewidth]{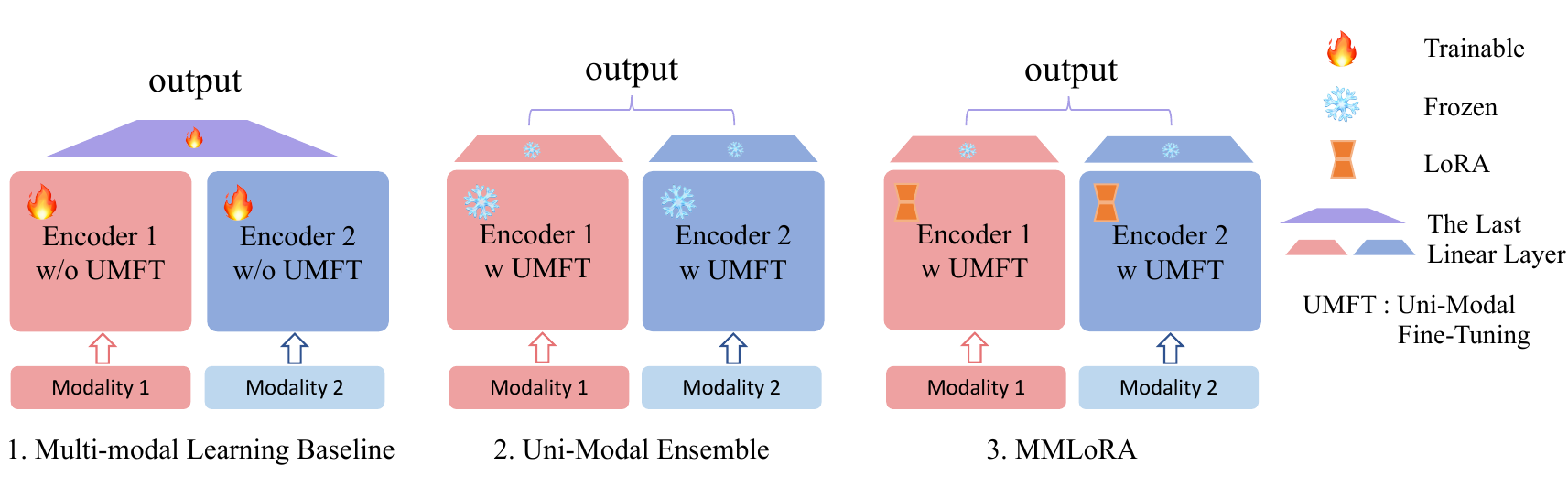}
    \caption{\textbf{Comparison of different multi-modal learning methods}. (1) The \textit{baseline method} directly uses large-scale pre-trained encoders to extract features of the corresponding modality. Then, it concatenates the features from different modalities and passes them through a linear layer to obtain the final prediction; (2) \textit{Uni-Modal Ensemble}~\citep{du2023uni} first fine-tunes the respective models with uni-modal data separately, and then directly averages the outputs from different modalities to obtain the final prediction; (3)  \textit{MMLoRA} first freezes the uni-modal fine-tuned models and introduces extra trainable rank decomposition matrices to \textit{a specific modality or all modalities' models}. It then performs multi-modal joint training, allowing the uni-modal fine-tuned models to better adapt across modalities, further enhancing the overall performance. }
    \label{fig:multi-modal learning}
\end{figure}

\textbf{Insufficient learning of uni-modal features in multi-modal learning.}
In multi-modal joint training, we use pre-trained models to encode their respective modalities, obtaining their features. We then concatenate the features from different modalities and pass them through a linear layer to produce a prediction. Through end-to-end training, we obtain a multi-modal model. The process is illustrated in the leftmost subfigure of Figure~\ref{fig:multi-modal learning}.
This approach is also referred to as late-fusion learning baseline and is widely used in multi-modal learning~\citep{du2023uni}. 
Its advantage is its applicability to a wide variety of different encoders. Then we employ linear evaluation~\citep{chen2020simple}, which trains a linear layer on frozen encoders, to assess their feature extraction ability. 
As Table~\ref{tab:laziness} shows, all encoders from multi-modal training are worse than their uni-modal fine-tuned counterparts.
\textit{Multi-modal learning with large-scale pre-trained encoders still suffer from insufficient learning of uni-modal features.}
This result is consistent with the scenario when using backbones that are either not pre-trained or pre-trained on ImageNet~\citep{peng2022balanced}, which has been proven to impact the model's generalization ability~\citep{du2023uni}.

In the following subsection, we introduce our proposed method to address this issue.

% \begin{table}[t]
%     \centering
%     \caption{Top-1 test accuracy (in \%) of linear evaluation on encoders from multi-modal training baseline and uni-modal
% fine-tuned models on AVE, Kinetics-Sound and CREMA-D. The one with better performance is highlighted in \textbf{bold}.}
%     \vspace{3pt}

%     \begin{tabular}{c c c c c c c}
% 		\toprule
% 		\multirow{2}*{\textbf{Methods.}} & \multicolumn{2}{c}{\textbf{AVE}} & \multicolumn{2}{c}{\textbf{Kinetics-Sound}}& \multicolumn{2}{c}{\textbf{CREMA-D}}\\
% 		\cline{2-7} & RGB & Audio & RGB & Audio& RGB & Audio\\
% 		\midrule
% 		  Multi-Modal Training & 83.3 &  79.8 & 78.7 & 67.9 & 70.3 & 64.8 \\
% 	    % \cdashline{1-7}
%             Uni-Modal Fine-Tuned & \textbf{88.1}  & \textbf{85.6} & \textbf{84.3} & \textbf{69.6} & \textbf{77.7} & \textbf{75.8}\\		
% 		\bottomrule
% \end{tabular}
%     \label{tab:laziness}
% \end{table}

\subsection{\textbf{M}ulti-\textbf{M}odal \textbf{Lo}w-\textbf{R}ank \textbf{A}daptation learning~(\textbf{MMLoRA})} 
\label{sec3_2_mmlora}

Multi-modal joint learning struggles to fully capture uni-modal features, and there have been previous papers attempting to address this issue~\citep{du2023uni,fan2023pmr,peng2022balanced}.
However, whether by introducing new loss functions~\citep{fan2023pmr} or dynamically balancing the learning progress of different modalities~\citep{peng2022balanced}, none can enable the encoders from the multi-modal training to achieve performance equivalent to that from uni-modal training.
\citet{du2023uni} points out that directly training uni-modal models separately and then averaging the outputs of the uni-modal models is already quite strong~(Uni-Modal Ensemble, as shown in Figure~\ref{fig:multi-modal learning}). However, \citep{du2023uni} also points out that UME is not suitable for all scenarios, as it cannot learn the so-called \textit{paired features}. As can be seen from Section 3.2 of the paper~\citep{du2023uni}, in multi-modal datasets, different modalities need to adapt to each other to produce better results.

% \textbf{Drawbacks of Uni-Modal Ensemble~(UME).} While UME is simple and effective, it does not engage in cross-modal interaction and adaptation. As can be seen from Section 3.2 of the paper~\citep{du2023uni}, in multi-modal datasets, different modalities need to adapt to each other to produce better results.

\textbf{MMLoRA.} Uni-Modal Fine-tuned models need further adaptation to each other to achieve better results. Given that uni-modal fine-tuned models have already learned a significant amount of features, our hypothesis is that only a small portion of parameters is needed for this adaptation. To this end, we draw inspiration from Parameter-Efficient Fine-Tuning~(PEFT), particularly from LoRA~\citep{hu2021lora} and propose \textbf{M}ulti-\textbf{M}odal \textbf{Lo}w-\textbf{R}ank \textbf{A}daptation learning~(\textbf{MMLoRA}). Specifically, we freeze the weights of the uni-modal fine-tuned models and introduce additional trainable rank decomposition matrices to a specific modality or all modalities' models. 
We then utilize multi-modal joint training to train these new parameters, enabling the modalities to better adapt to each other. 
Specifically, we average the uni-modal predictions to obtain a multi-modal prediction, then compute the loss against the label, from which we derive the gradients to update the LoRA parameters. 
We also illustrate MMLoRA in Figure~\ref{fig:multi-modal learning}.

\textbf{Formal Definition of MMLoRA.} We suppose there are $M$ modalities within the problem. For each single modality $m=1, ..., M$, we have a dataset $\mathcal{Z}^{m}=\left\{\left(x_j^{m}, y_j\right)\right\}_{j=1, \ldots, N}$ containing the training data pairs of input~($x^m$) and label~($y$). The datasets with all modalities are denoted as $\bar{\mathcal{Z}}$. Then we denote the uni-modal model for each single modality with its weight $\Phi^{m}$ as $P_{\Phi^{m}}(y|x^{m})$, which will be short as $P^{m}_{\Phi}(y|x)$ for simplicity. 
We first apply Uni-Modal Fine-Tuning~(UMFT) by optimizing each encoder with its own uni-modal data separately with fully fine-tuning, which updates the whole model by repeatedly following the gradient of 
%which update the model from $\Phi_0^{m}$ for the encoder in single modality $m$, by some $\Delta \Phi^m$ obtained from repeatedly updating following the gradient of

\begin{equation}\label{eq:UMFT}
\max _{\Phi^{m}} \sum_{(x, y) \in \mathcal{Z}^{m}} \log \left(P^{m}_{\Phi}\left(y| x\right)\right), m=1, ..., M.
\end{equation}
% Hereby, we obtain the uni-modal models after UMFT with weight $\Phi_\text{UMFT}^{m}$, $m=1, ..., M$.

 Recalling the reparameterization scheme of LoRA~\citep{hu2021lora}, for a pre-trained weight matrix $W_0 \in \mathbb{R}^{d \times k}$, the parameter adaption $\Delta W$ can be constrained in a low-rank decomposition as
\begin{equation}\label{eq:lora_update}
W_0 +\Delta W =W_0+B A,
\end{equation}
where $A\in \mathbb{R}^{r \times k}$ and $B\in \mathbb{R}^{d \times r}$ are matrices with rank $r \ll \min (d, k)$. $A$ is Gaussian initialized and $B$ is set as  azero matrix to ensure the update $\Delta W $ is zero at the beginning of training.

In MMLoRA, we apply the LoRA reparameterization scheme (\ref{eq:lora_update}) to  uni-modal fine-tuned models $\Phi_\text{UMFT}^{m}$ obtained by Eq.~\ref{eq:UMFT}.
For each uni-modal fine-tuned model reparametrized by LoRA, the real trainable parameters $\Theta^m$ are significantly fewer than the total parameters in $\Phi^m$, inheriting the advantage of LoRA in parameter efficiency and preventing overfitting. 
Once training is completed, these newly introduced parameters can be directly merged with the original parameters~(as Eq.~\ref{eq:lora_update} shows).
We denote the LoRA update of each uni-modal model as $\Delta \Phi^m = \Delta \Phi(\Theta^m)$. 
Empirically, we have found that simply allowing only one modality to be reparameterized by LoRA, and then updating this modality's LoRA parameters to adapt another modality through multi-modal joint learning, is also very effective. To this end, we have the flexibility to selectively update specific $\Theta^m$. We denote the collection of selected indexes $m$ of each modality as the set $\mathcal{M}$. The selected parameters $\{\Theta^m\}_{m\in \mathcal{M}}$ to be updated are collectively represented as $\Theta$. Then we optimize the multi-modal joint training objective (\ref{eq:mmlora_update}) towards $\Theta$ with the entire multi-modal dataset $\bar{\mathcal{Z}}$ as

\begin{equation}\label{eq:mmlora_update}
\max_{\Theta} \sum_{(x, y) \in \bar{\mathcal{Z}}}  \log \sum_{m=1}^M P^m_{\Phi_\text{UMFT}+\Delta \Phi(\Theta)}\left(y|x\right).
\end{equation}
Finally, we obtain the MMLoRA model as $P^{\text{MMLoRA}}\left(y|x\right) = \frac{1}{M} \sum\limits_{m=1}^M P^m_{\Phi_\text{UMFT}+\Delta \Phi(\Theta)}\left(y|x\right)$, where $P^m_{\Phi_\text{UMFT}+\Delta \Phi(\Theta)}$ denotes a uni-modal model firstly trained by Uni-Modal Fine-Tuning~(UMFT, \ref{eq:UMFT}), then reparameterized through LoRA, followed by a multi-modal joint training (\ref{eq:mmlora_update}). %The MMLoRA allows different modalities to adapt to each other more effectively, ultimately leading to improved performance.

% Note that, empirically, applying LoRA reparametrization to just one uni-modal model and then conducting multi-modal joint training, allowing this modality's model to adapt to the other modality through LoRA parameters update, is also effective. 

Overall, MMLoRA introduces a limited number of new parameters for the uni-modal fine-tuned models, which provides an opportunity for different modalities to adapt to each other more effectively, ultimately leading to improved performance. 

In the following section, we will demonstrate the effectiveness of MMLoRA and conduct ablation experiments to gain a deeper understanding of the underlying principles that drive its success.

\section{Experiment}
\label{sec4}
% In this section, we first introduce the datasets we utilize, the pre-trained models we employ, and settings for various other relevant parameters. 
In this section, we begin by describing the six multi-modal datasets~(including AVE, Kinetics-Sound, CREMA-D, MM-IMDB, UPMC Food101 and UCF101) we use, pre-trained models we select, and other settings.
Subsequently, we present our primary experimental results, demonstrating the significant performance improvement by using pre-trained large models and the effectiveness of MMLoRA across a range of multi-modal datasets. 
Lastly, we conduct ablation studies on MMLoRA to further understand the underlying mechanisms of its operation.

\subsection{Datasets and Experimental Settings}

\subsubsection{Datasets} 
\label{sec4_data}
% We run our experiments on three categories of datasets: audio-visual (e.g., AVE, Kinetics-Sound, CREMA-D), vision-language (e.g., MM-IMDB, UPMC Food-101), and RGB-Optical Flow (UCF101), totaling six multi-modal datasets. Below, we delve into the specific details of these datasets.

\textbf{1. Audio-Visual Datasets:}
(1) The \textit{Audio-Visual Event localization~(AVE) dataset}, as introduced in~\citet{tian2018audio}, contains 4,143 unrestricted 10-second videos across 28 event types such as Rodents, Accordion, Mandolin and so on. AVE is a subset of AudioSet~\citep{gemmeke2017audio}. The train/val/test splits are 3,339/402/402 videos, respectively;
(2) The \textit{Kinetics-Sounds} dataset, referenced in~\citep{arandjelovic2017look}, is a curated subset of Kinetics400, which features YouTube videos with hand-labeled human actions. This subset encompasses 32 classes, including actions like playing the harmonica, tapping a pen, shoveling snow and so on. The dataset is divided into 22,728 training videos and 1,593 validation videos.
(3) The \textit{CREMA-D} dataset~\citep{cao2014crema}, is an audio-visual dataset designed for speech emotion recognition. It features 7,442 short video clips (2-3 seconds each) from 91 actors expressing six common emotions: angry, happy, sad, neutral, discarding, disgust, and fear. Emotion labels were sourced from 2,443 crowd-sourced raters. The dataset is split into a 6,698-sample training set and a validation set in a 9:1 ratio, with a separate 744-sample test set.

\textbf{2. Vision-Language Datasets:}
(1) The \textit{MM-IMDB} dataset, introduced by \citet{arevalo2017gated}, combines movie plot outlines with movie posters for genre classification. Each movie may belong to multiple genres, making it a multilabel prediction task. The dataset was created to address the limited availability of quality multi-modal classification datasets. The train/val/test splits are 15,552/2,608/7,799 videos, respectively;
(2) The UPMC FOOD101 dataset, introduced by \citet{wang2015recipe}, offers textual descriptions of recipes spanning 101 food categories. These descriptions, extracted from curated web pages, are paired with an image sourced from Google Image Search, which might occasionally match a noisy category. The goal is to determine the appropriate food label for each text-image pair. The train/test splits are 67,972/22,716 videos, respectively.

\textbf{3. RGB-Optical Flow Dataset, UCF101.}
The UCF101 dataset, presented by \citet{soomro2012ucf101}, is designed for action recognition, featuring 101 distinct action categories. It contains around 7k training videos and 3k testing videos. For our experiments, we employ the RGB and flow data supplied by ~\citet{feichtenhofer2016convolutional}.

\subsubsection{Experimental Settings}
\label{sec4_encoder_select}
\textbf{Selection of Pre-Trained Encoders.} 
In Table~\ref{tab:encoder_select}, we display the encoders we selected for different datasets. On the audio-visual dataset, we employ ViT-B as the visual encoder, as it is already quite effective, outperforming previous multi-modal models. 
Although CLIP has trained a text encoder aligned with images, its text encoder is not as effective as those directly pre-trained on text~\citep{saharia2022photorealistic}. 
Therefore, we choose for text encoders that were purely pre-trained on text, such as BERT and DeBERTa.
For UPMC Food101, we select BERT-Base as various text encoders show similar results.
To better demonstrate the effectiveness of MMLoRA in comparison to the baseline methods, we also implement MMLoRA using the same encoders as the baselines for a direct comparison. In such cases, we will point it out.
Additionally, we discover that bigger models are not necessarily better. We've included more details in the Appendix~\ref{appendix_bigger}.

\textbf{Data Preprocessing.}
(1) For \textit{images}, we randomly resize to 224x224, then apply horizontal flip, and normalize them using the OpenCLIP MEAN and STD values. In audio-visual datasets, we randomly take 3 frames as input when training, and put them into the 2D network as \citet{peng2022balanced} does;
(2) For \textit{audio}, we first ensure a sample rate of 16k, then convert the t-second audio waveform into 128-dimensional log Mel filterbank sequences. Our audio preprocessing is consistent with that of \citet{girdhar2023imagebind} or \citet{gong2021ast};
(3) For \textit{text}, we use the corresponding tokenizer for different language models. Additionally, we set the maximum token sequence length to 512.
% Specifically, BERT-base uses the \textit{bert-base-uncased} tokenizer, while DeBERTa-Large employs the \textit{microsoft/deberta-large} tokenizer. 

\textbf{Optimizer.}
When fine-tuning the language model in a uni-modal setting, we use BertAdam~\citep{devlin2018bert} as the optimizer; We use SGD when training the optical flow in UCF101; For all other experiments, we employ AdamW~\citep{loshchilov2017decoupled}. For the Vision-language task, we use a batch size of 160, while for other tasks, we use a batch size of 64.

\textbf{MMLoRA Settings.} Unless oterwise specified, we set the rank of LoRA to 1. For transformer~\citep{vaswani2017attention} models, we add LoRA to all Query, Key, and Value layers. For ResNet~\citep{he2016deep}, we apply LoRA to all convolutional layers. For the CREMA-D dataset, we only reparametrize the audio model using LoRA. For UCF101, only the optical flow model is reparametrized with LoRA. In other cases, the results we report for MMLoRA involve reparametrization using LoRA for all modalities. We further conduct an ablation study on this aspect in Sec~\ref{sec4_lora_which_part}. 

\textbf{Other Hyper-parameters.}
In this sub-section, we've outlined some common settings. There are other settings, such as the learning rate, that vary by task. Due to space constraints, we have placed other hyper-parameters in Appendix~\ref{appendix_settings}.

\subsection{Main experimental results}

\begin{table}[t]
    \centering
    \caption{\textbf{Comparison of linear evaluation and fully fine-tuning} of pre-trained models on uni-modal data of AVE, Kinetics-Sound~(KS), CREMA-D, MM-IMDB and UPMC Food101. The reported evaluation metric are Top-1 Accuracy (AVE, Kinetics-Sound , CREMA-D and UPMC Food101) and F1-Micro (MM-IMDB).
    Better performance is highlighted in \textbf{bold}.}
    \vspace{3pt}

    \begin{tabular}{c c c |c c |c c |c c |c c}
		\toprule
		\multirow{2}*{\textbf{Method}} & \multicolumn{2}{c}{\textbf{AVE}} & \multicolumn{2}{c}{\textbf{KS}}& \multicolumn{2}{c}{\textbf{CREMA-D}}& \multicolumn{2}{c}{\textbf{MM-IMDB}}& \multicolumn{2}{c}{\textbf{Food101}}\\
		\cline{2-11} & RGB & Audio & RGB & Audio& RGB & Audio & RGB & Text& RGB & Text\\
		\midrule
		  linear eval & 87.6 &  82.1 & 79.0 & 66.2 & 51.2 & 60.6 &48.7 & 22.1 &81.8 & 21.2 \\
	    % \cdashline{1-7}
            fine-tuning & \textbf{88.1}  & \textbf{85.6} & \textbf{84.3} & \textbf{69.6} & \textbf{77.7} & \textbf{75.8} & \textbf{60.2} & \textbf{68.6} & \textbf{84.3} & \textbf{86.6}\\		
		\bottomrule
\end{tabular}
    \label{tab:linear_finetune}
\end{table}

\subsubsection{For large-scale pre-trained models: directly use their features or perform fine-tuning?}
Large-scale pre-trained models have proven to exhibit strong zero-shot performance~\citep{radford2021learning}. However, it seems that only when both the data volume and model size reach a considerably large scale does zero-shot outperform fine-tuning~\citep{openai2023gpt4}.
In this sub-section, we first compare two methods using our data: the first approach involves adding a linear layer to the pre-trained models and training only this layer; the second approach entails fully fine-tuning the models. 
Note that for this experiment, we train each model only on uni-modal data. 
% The learning rates used are: \(1 \times 10^{-4}\)~(visual models in audio-visual datasets),  \(1 \times 10^{-5}\)~(audio models in audio-visual datasets, visual models in vision-language datasets) and \(5 \times 10^{-5}\)~(language models).
The results are shown in the Table~\ref{tab:linear_finetune}. Fine-tuning pre-trained models on uni-modal data significantly outperforms linear evaluation.
Thus, in our subsequent experiments, we won't directly use features extracted from the pre-trained model; instead, we will train the encoeder's parameters with our data.

\subsubsection{The effectiveness of Large-scale pre-trained models}
\label{sec_4_effect_largemodels}
We present the results of different methods using various backbones on different datasets in Tables \ref{tab:mmlora_av_dataset}, \ref{tab:mmlora_vl}, and \ref{tab:mmlora_ucf}. In these three tables, the results below the dashed line use stronger large-scale pre-trained encoders than those above. Notably, especially in Table \ref{tab:mmlora_av_dataset} and \ref{tab:mmlora_ucf}, the performance of the methods below the dashed line far surpasses that of the methods above. In Table~\ref{tab:mmlora_vl}, with improved backbones, MMLoRA's performance is also significantly enhanced.
The evident performance gap highlights the necessity of using large-scale pre-trained models.
% This performance gap cannot be bridged by mere algorithmic design. These comparisons unquestionably demonstrate the necessity of using large-scale pre-trained models.

\subsubsection{Main Results of MMLoRA}
In this section, we demonstrate the effectiveness of MMLoRA across three different types of datasets and various backbones.% Before that, we revisit the implementation approach of MMLoRA: 
% First, we fine-tune the aforementioned pre-trained models on uni-modal data. 
% Then, we reparameterize the uni-modal fine-tuned models using LoRA. 
% By calculating the average outputs of the two uni-modal models and computing the loss with the label, we derive the gradients and update the trainable parameters. 

% For the CREMA-D dataset, we only reparametrize the audio model using LoRA; For UCF101, we only reparametrize the visual model using LoRA, and we conduct an ablation study on this in Sec~\ref{sec4_3_ablation}.

\begin{table}[t]
    \centering
    \caption{\textbf{Top-1 Test Accuracy of different methods on Audio-Visual Datasets}~(AVE, Kinetics-Sound and CREMA-D). \textit{Avg Acc} represents the average accuracy across the three datasets. * indicates our implementation.
    The best performance under same backbones is highlighted in \textbf{bold}.} 
    \vspace{3pt}
    \begin{tabular}{c|c|c|c|c|c}
    \toprule
       \textbf{Method} & \textbf{Backbone (A/V)} & \textbf{AVE} & \textbf{KS} & \textbf{CREMA-D} & \textbf{Avg Acc.}  \\
    \midrule
     G-Blending~\citep{wang2020makes}    & ResNet18/ResNet18 & 65.5& 62.2& 58.7 & 62.1 \\
     OGM-GE~\citep{peng2022balanced} & ResNet18/ResNet18 & 76.9 & 63.1 & 62.2 & 67.4 \\
     PMR~\citep{fan2023pmr} & ResNet18/ResNet18 & 74.3 & -&65.3 &-\\
     UME*~\citep{du2023uni} & ResNet18/ResNet18 & 85.4 & 78.8 & 78.2& 80.8\\
     \textbf{MMLoRA}~(ours)  & ResNet18/ResNet18 & \textbf{86.9} & \textbf{79.4}& \textbf{81.9}& \textbf{82.7}\\
     \hdashline
     Multi-Modal Baseline* & ViT-B/ViT-B & 94.7 & 90.6 & 87.6 & 91.0\\
     Classifier on frozen features* & ViT-B/ViT-B & 93.7 & 90.1 & 85.3 & 89.7 \\
     MBT~\citep{nagrani2021attention} & ViT-B/ViT-B & - & 85.0 & - & - \\
     UMT*~\citep{du2023uni} & ViT-B/ViT-B & 93.7  & 90.3  & 87.8 & 90.6 \\
     OGM-GE*~\citep{peng2022balanced} & ViT-B/ViT-B & 95.5 & 90.4 & 88.4 & 91.4 \\
     UME*~\citep{du2023uni} & ViT-B/ViT-B & 95.4  & 90.8 & 87.8 & 91.3 \\
     Fully Fine-tuned UME* & ViT-B/ViT-B & 95.2 & 91.3 & 87.5 & 91.3 \\
     \textbf{MMLoRA}~(ours) & ViT-B/ViT-B & \textbf{96.2}  & \textbf{91.4} & \textbf{88.6} & \textbf{92.1} \\
    \bottomrule
    \end{tabular}
    \label{tab:mmlora_av_dataset}
\end{table}

\begin{figure}[t]
\begin{minipage}[t]{\textwidth}
\begin{minipage}[t]{0.525\textwidth}
\makeatletter\def\@captype{table}
\centering

\caption{\textbf{The performance of MMLoRA and other methods on UPMC Food101~(Acc.) and MM-IMDB~(F1-Micro/F1-Macro)}. Above the dashed line, the backbone used is consistent with \citet{kiela2019supervised}, while below the dashed line, the methods compared are reimplemented using improved backbone by \citet{li2023efficient}.}
\vspace{3pt}
\begin{tabular}{l|c|c}
    \toprule
    \textbf{Method}     & \textbf{Food101} & \textbf{MM-IMDB}      \\
    \midrule
    MMBT & 92.1 & 66.8/61.6\\
    \textbf{MMLoRA}~(ours) & \textbf{93.7} & \textbf{67.2/61.7} \\
    \hdashline
    Baseline & 93.29&64.9/59.6\\
    PMF  & 91.51 & 64.5/58.8\\
    PMF-L & 91.68 & 66.7/61.7\\
    MBT & 93.6 & 64.8/59.6 \\
    MMBT & 94.10 & 66.1/60.8\\
    \textbf{MMLoRA}~(ours) &\textbf{95.9} &\textbf{71.7/67.5}\\
    \bottomrule
\end{tabular}

\label{tab:mmlora_vl}
\end{minipage}
    \hspace{15pt}
\begin{minipage}[t]{0.425\textwidth}
\makeatletter\def\@captype{table}
\centering
\caption{\textbf{Top-1 Test Accuracy (in \%) of different methods on UCF101.} * indicates our implementation and the performance of the other methods is derived from \citet{du2023uni}. We also present the backbones used for different methods (RGB/Optical Flow). }
\vspace{3pt}
\begin{tabular}{l|c|c}
    \toprule
    \textbf{Method}  &   \textbf{Backbone} & \textbf{Acc.}\\
    \midrule
    MM Baseline & Res18/Res18 & 82.3  \\
    G-Blending & Res18/Res18  & 83.0 \\
    OGM-GE & Res18/Res18 & 84.0 \\
    UMT & Res18/Res18 & 84.5\\
    UME & Res18/Res18 & 86.8\\
    \textbf{MMLoRA} & Res18/Res18 & \textbf{87.1} \\
    \hdashline
    UME* & ViT-B/Res18 & 93.0\\
    \textbf{MMLoRA} & ViT-B/Res18 & \textbf{93.4}\\
    
    \bottomrule
\end{tabular}
\label{tab:mmlora_ucf}

\end{minipage}
\end{minipage}
\end{figure}

\begin{table}[t]
    \centering
    \caption{\textbf{Top-1 Test Accuracy (in \%) of linear evaluation on encoders from MMLoRA} and uni-modal
fine-tuned models on AVE, Kinetics-Sound and CREMA-D. In this experiment, the encoders from MMLoRA are all re-parameterized with $r=1$ and then performed multi-modal joint training. }
    \vspace{3pt}

    \begin{tabular}{c c c| c c| c c}
		\toprule
		\multirow{2}*{\textbf{Encoder Source}} & \multicolumn{2}{c}{\textbf{AVE}} & \multicolumn{2}{c}{\textbf{Kinetics-Sound}}& \multicolumn{2}{c}{\textbf{CREMA-D}}\\
		\cline{2-7} & RGB & Audio & RGB & Audio& RGB & Audio\\
		\midrule
		  Uni-Modal Fine-Tuned & 88.1  & 85.6 & 84.3 & 69.6 & 77.7 & \textbf{75.8}\\
	    % \cdashline{1-7}
            \textbf{MMLoRA}~(ours) & 88.1 & \textbf{86.4} & \textbf{85.3}& \textbf{69.8}& \textbf{78.1} & 75.7\\		
		\bottomrule
\end{tabular}
    \label{tab:mmlora_unimodal}
\end{table}

\textbf{MMLoRA on audio-visual datasets.}
As Table~\ref{tab:mmlora_av_dataset} shown, we can observe that MMLoRA consistently exhibits the best performance across different backbones.
The methods compared include balancing training by adding loss~(G-Blending~\citep{wang2020makes}, PMR~\citep{fan2023pmr}, UMT~\citep{du2023uni}), adjusting the training progress of different modalities by modifying gradients~(OGM-GE~\citep{peng2022balanced}), novel fusion framework~(MBT~\citep{nagrani2021attention}),or directly averaging uni-modal predictions~(UME~\citep{du2023uni}). We also compare multi-modal joint full fine-tuning of the uni-modal fine-tuned models~(namely \textit{Fully Fine-tuned UME}), and the results are inferior to MMLoRA. We hypothesize that the limited data, combined with the excessive model parameters, makes full fine-tuning less effective than parameter-efficient fine-tuning. \textit{Classifier on frozen features} refers to using uni-modal fine-tuned models to extract features and training a multi-modal linear layer.
And the multi-modal baseline method is illustrated in the leftmost subfigure of Figure~\ref{fig:multi-modal learning}.

\textbf{MMLoRA on vision-language datasets.}\label{MMLoRA_vldata}
Here, we compare MMLoRA with MBT~\citep{nagrani2021attention}, MMBT~\citep{kiela2019supervised}, and PMF~\citep{li2023efficient}. When we use the same backbone as the original MMBT paper, MMLoRA outperforms MMBT. 
Additionally, when we employ a superior backbone, MMLoRA also surpasses various methods reimplemented by \citet{li2023efficient}.

% \begin{table}[t]
%     \centering
%     \caption{\textbf{Top-1 Test Accuracy of different methods on Audio-Visual Datasets}~(AVE, Kinetics-Sound and CREMA-D). * indicates our implementation.
%     The best performance under same backbones is highlighted in \textbf{bold}.} 
%     \vspace{3pt}
%     \begin{tabular}{c|c|c|c|c|c}
%     \toprule
%        Method & Backbone(A/V) & AVE & KS & CREMA-D & Avg Acc  \\
%     \midrule
%      G-Blending~\citep{wang2020makes}    & ResNet18/ResNet18 & 65.5& 62.2& 58.7 & 62.1 \\
%      OGM-GE~\citep{peng2022balanced} & ResNet18/ResNet18 & 76.9 & 63.1 & 62.2 & 67.4 \\
%      PMR~\citep{fan2023pmr} & ResNet18/ResNet18 & 74.3 & -&65.3 &-\\
%      UME*~\citep{du2023uni} & ResNet18/ResNet18 & 85.4 & 78.8 & 78.2& 80.8\\
%      MMLoRA~(ours)  & ResNet18/ResNet18 & \textbf{86.9} & \textbf{79.4}& \textbf{81.9}& \textbf{82.7}\\
%      \hdashline
%      Multi-Modal Baseline* & ViT-B/ViT-B & 94.7 & 90.6 & 87.6 & 91.0\\
%      Classifier on frozen features* & ViT-B/ViT-B & 93.7 & 90.1 & 85.3 & 89.7 \\
%      OGM-GE*~\citep{peng2022balanced} & ViT-B/ViT-B & 95.5 & 90.4 & 88.4 & 91.4 \\
%      UME*~\citep{du2023uni} & ViT-B/ViT-B & 95.4  & 90.8 & 87.8 & 91.3 \\
%      MMLoRA~(ours) & ViT-B/ViT-B & \textbf{96.2}  & \textbf{91.4} & \textbf{88.6} & \textbf{92.1} \\
%     \bottomrule
%     \end{tabular}
%     \label{tab:mmlora_av_dataset}
% \end{table}

\textbf{MMLoRA on UCF101.}
The Uni-Modal Ensemble~\citep{du2023uni} has already proven to be very effective on UCF101. As Table~\ref{tab:mmlora_ucf} shows, we apply MMLoRA to UCF101 to further boost performance, regardless of the backbone used. 
% The methods also compared here include G-Blending~\citep{wang2020makes} and OGM-GE~\citep{peng2022balanced}.
% and their performance is primarily reported in \citet{du2023uni}.

\subsection{Ablation Study on MMLoRA}
\label{sec4_3_ablation}

\textbf{MMLoRA does not affect the uni-modal feature extraction.}
MMLoRA introduces new trainable parameters to the uni-modal fine-tuned models and conducts multi-modal joint training. Taking the encoders trained by MMLoRA for linear evaluation, as shown in Table~\ref{tab:mmlora_unimodal}, not only does MMLoRA not affect the extraction of uni-modal features, but in some cases, it even surpasses its uni-modal training counterpart.

\textbf{Apply LoRA re-parameterizion to which part?} 
\label{sec4_lora_which_part}
In Table~\ref{tab:lora_encoder}, we experiment with adding LoRA to just one uni-modal fine-tuned model~~(during this process, we essentially adjust only the LoRA parameters of a particular modality to make it adapt to another modality.) and compare it to that of applying LoRA to both models. 
Firstly, we can see that all of the resulting models always outperform Uni-Modal Ensemble (Refer to the Table~\ref{tab:mmlora_av_dataset}).
On the CREMA-D dataset, applying LoRA solely to the Audio Model, followed by multi-modal joint training, yields better results. 
And for the other two datasets, directly applying LoRA to both modalities is superior. We hypothesize that allowing only one modality to update its  parameters to adapt to the other modality might lead to more stable training in some scenarios.

\begin{table}[t]
    \centering
    \caption{\textbf{Comparison between the MMLoRA model where LoRA is applied to a single modality and the MMLoRA model where LoRA is applied to all modalities} on AVE, Kinetics-Sound and CREMA-D. \textit{RGB} or \textit{Audio} indicates that we only re-parameterize the fine-tuned RGB or Audio Model by LoRA, and then, through multi-modal joint training, we allow the specified modality to adapt to the other. \textit{Both} means that we apply LoRA re-parameterization to both modalities and then jointly train them. The one with better performance is highlighted in \textbf{bold}. }
    \vspace{3pt}

    \begin{tabular}{c c c c |c c c| c c c}
		\toprule
		\multirow{2}*{\textbf{Method}} & \multicolumn{3}{c}{\textbf{AVE}} & \multicolumn{3}{c}{\textbf{Kinetics-Sound}}& \multicolumn{3}{c}{\textbf{CREMA-D}}\\
		\cline{2-10} & RGB & Audio & Both & RGB& Audio & Both & RGB & Audio & Both\\
		\midrule

            \textbf{MMLoRA} & 95.7 & \textbf{96.2} & \textbf{96.2}&91.3 & 91.3 & \textbf{91.4} & 87.9 & \textbf{88.6}& 88.3\\		
		\bottomrule
\end{tabular}
    \label{tab:lora_encoder}
\end{table}

\begin{table}[t]
    \centering
    \caption{\textbf{Top-1 Test Accuracy (in \%) of MMLoRA with different ranks} on the Kinetics-Sound and AVE datasets. We include the results of UME here for comparison.}
    \vspace{3pt}
    \begin{tabular}{c|c|c|c|c|c|c}
    \toprule
       \diagbox{Dataset}{Rank ($r$)}  & \textcolor{gray}{UME}& \quad 1 \quad\quad & \quad 2 \quad\quad & \quad 4 \quad\quad & \quad 8 \quad\quad & \quad 64 \quad\quad  \\
    \midrule
     AVE    & \textcolor{gray}{95.4} & 96.2& \textbf{96.5}& \textbf{96.5} & 96.2 &  95.2\\
     \midrule
      Kinetics-Sound & \textcolor{gray}{90.8}   & \textbf{91.4} & 91.3 &\textbf{91.4} & 91.3 & 91.1 \\
      
    \bottomrule
    \end{tabular}
    \label{tab:mmlora_rank}
\end{table}

\textbf{Rank Selection in MMLoRA.}
We also conduct an ablation study on the rank values of MMLoRA on the Kinetics-Sound and AVE datasets.
As shown in Table~\ref{tab:mmlora_rank}, when the rank is set too high, the performance declines~(in AVE, the performance of MMLoRA with a rank of 64 might even be inferior to UME). This might be due to the introduction of too many parameters while the dataset isn't large enough.
Although setting the rank to 1 doesn't always yield the best results, the performance is consistently good. These results suggests that cross-modal adaptation is necessary and might not require many parameters, and thus, in our main experiments with MMLoRA, we set the rank to 1.

\textbf{Other Ablation Study.} We try reparametrizing pre-trained models with LoRA directly and then conduct multi-modal joint training. We also conducted an ablation study on whether to add LoRA to the final linear layer. Results are in the Appendix~\ref{appendix_ablation} due to space constraints.
\section{Conclusion}
Large-scale pre-trained models have undoubtedly significantly enhanced discriminative multi-modal learning, yet they also encounter the issue of sufficient learning of uni-modal features. We introduce \textbf{M}ulti-\textbf{M}odal \textbf{Lo}w-\textbf{R}ank \textbf{A}daptation learning (\textbf{MMLoRA}), which adds a small amount of parameters to uni-modal fine-tuned models and then engages in multi-modal joint training to better adapt across modalities, thereby boosting the overall performance. We hope this paper can bring some new insights to the field of discriminative multi-modal learning.

\bibliography{iclr2024_conference}

\begin{thebibliography}{51}
\providecommand{\natexlab}[1]{#1}
\providecommand{\url}[1]{\texttt{#1}}
\expandafter\ifx\csname urlstyle\endcsname\relax
  \providecommand{\doi}[1]{doi: #1}\else
  \providecommand{\doi}{doi: \begingroup \urlstyle{rm}\Url}\fi

\bibitem[Anil et~al.(2023)Anil, Dai, Firat, Johnson, Lepikhin, Passos, Shakeri,
  Taropa, Bailey, Chen, et~al.]{anil2023palm}
Rohan Anil, Andrew~M Dai, Orhan Firat, Melvin Johnson, Dmitry Lepikhin,
  Alexandre Passos, Siamak Shakeri, Emanuel Taropa, Paige Bailey, Zhifeng Chen,
  et~al.
\newblock Palm 2 technical report.
\newblock \emph{arXiv preprint arXiv:2305.10403}, 2023.

\bibitem[Arandjelovic \& Zisserman(2017)Arandjelovic and
  Zisserman]{arandjelovic2017look}
Relja Arandjelovic and Andrew Zisserman.
\newblock Look, listen and learn.
\newblock In \emph{Proceedings of the IEEE international conference on computer
  vision}, pp.\  609--617, 2017.

\bibitem[Arevalo et~al.(2017)Arevalo, Solorio, Montes-y G{\'o}mez, and
  Gonz{\'a}lez]{arevalo2017gated}
John Arevalo, Thamar Solorio, Manuel Montes-y G{\'o}mez, and Fabio~A
  Gonz{\'a}lez.
\newblock Gated multimodal units for information fusion.
\newblock \emph{arXiv preprint arXiv:1702.01992}, 2017.

\bibitem[Brown et~al.(2020)Brown, Mann, Ryder, Subbiah, Kaplan, Dhariwal,
  Neelakantan, Shyam, Sastry, Askell, et~al.]{brown2020language}
Tom Brown, Benjamin Mann, Nick Ryder, Melanie Subbiah, Jared~D Kaplan, Prafulla
  Dhariwal, Arvind Neelakantan, Pranav Shyam, Girish Sastry, Amanda Askell,
  et~al.
\newblock Language models are few-shot learners.
\newblock \emph{Advances in neural information processing systems},
  33:\penalty0 1877--1901, 2020.

\bibitem[Cao et~al.(2014)Cao, Cooper, Keutmann, Gur, Nenkova, and
  Verma]{cao2014crema}
Houwei Cao, David~G Cooper, Michael~K Keutmann, Ruben~C Gur, Ani Nenkova, and
  Ragini Verma.
\newblock Crema-d: Crowd-sourced emotional multimodal actors dataset.
\newblock \emph{IEEE transactions on affective computing}, 5\penalty0
  (4):\penalty0 377--390, 2014.

\bibitem[Chen et~al.(2020{\natexlab{a}})Chen, Radford, Child, Wu, Jun, Luan,
  and Sutskever]{chen2020generative}
Mark Chen, Alec Radford, Rewon Child, Jeffrey Wu, Heewoo Jun, David Luan, and
  Ilya Sutskever.
\newblock Generative pretraining from pixels.
\newblock In \emph{International conference on machine learning}, pp.\
  1691--1703. PMLR, 2020{\natexlab{a}}.

\bibitem[Chen et~al.(2020{\natexlab{b}})Chen, Kornblith, Norouzi, and
  Hinton]{chen2020simple}
Ting Chen, Simon Kornblith, Mohammad Norouzi, and Geoffrey Hinton.
\newblock A simple framework for contrastive learning of visual
  representations.
\newblock In \emph{International conference on machine learning}, pp.\
  1597--1607. PMLR, 2020{\natexlab{b}}.

\bibitem[Cherti et~al.(2023)Cherti, Beaumont, Wightman, Wortsman, Ilharco,
  Gordon, Schuhmann, Schmidt, and Jitsev]{cherti2023reproducible}
Mehdi Cherti, Romain Beaumont, Ross Wightman, Mitchell Wortsman, Gabriel
  Ilharco, Cade Gordon, Christoph Schuhmann, Ludwig Schmidt, and Jenia Jitsev.
\newblock Reproducible scaling laws for contrastive language-image learning.
\newblock In \emph{Proceedings of the IEEE/CVF Conference on Computer Vision
  and Pattern Recognition}, pp.\  2818--2829, 2023.

\bibitem[Dai et~al.(2023)Dai, Li, Li, Tiong, Zhao, Wang, Li, Fung, and
  Hoi]{dai2023instructblip}
Wenliang Dai, Junnan Li, Dongxu Li, Anthony Meng~Huat Tiong, Junqi Zhao,
  Weisheng Wang, Boyang Li, Pascale Fung, and Steven Hoi.
\newblock Instructblip: Towards general-purpose vision-language models with
  instruction tuning, 2023.

\bibitem[Dettmers et~al.(2023)Dettmers, Pagnoni, Holtzman, and
  Zettlemoyer]{dettmers2023qlora}
Tim Dettmers, Artidoro Pagnoni, Ari Holtzman, and Luke Zettlemoyer.
\newblock Qlora: Efficient finetuning of quantized llms.
\newblock \emph{arXiv preprint arXiv:2305.14314}, 2023.

\bibitem[Devlin et~al.(2018)Devlin, Chang, Lee, and Toutanova]{devlin2018bert}
Jacob Devlin, Ming-Wei Chang, Kenton Lee, and Kristina Toutanova.
\newblock Bert: Pre-training of deep bidirectional transformers for language
  understanding.
\newblock \emph{arXiv preprint arXiv:1810.04805}, 2018.

\bibitem[Du et~al.(2023)Du, Teng, Li, Liu, Yuan, Wang, Yuan, and
  Zhao]{du2023uni}
Chenzhuang Du, Jiaye Teng, Tingle Li, Yichen Liu, Tianyuan Yuan, Yue Wang, Yang
  Yuan, and Hang Zhao.
\newblock On uni-modal feature learning in supervised multi-modal learning.
\newblock \emph{arXiv preprint arXiv:2305.01233}, 2023.

\bibitem[Fan et~al.(2023)Fan, Xu, Wang, Wang, and Guo]{fan2023pmr}
Yunfeng Fan, Wenchao Xu, Haozhao Wang, Junxiao Wang, and Song Guo.
\newblock Pmr: Prototypical modal rebalance for multimodal learning.
\newblock In \emph{Proceedings of the IEEE/CVF Conference on Computer Vision
  and Pattern Recognition}, pp.\  20029--20038, 2023.

\bibitem[Feichtenhofer et~al.(2016)Feichtenhofer, Pinz, and
  Zisserman]{feichtenhofer2016convolutional}
Christoph Feichtenhofer, Axel Pinz, and Andrew Zisserman.
\newblock Convolutional two-stream network fusion for video action recognition.
\newblock In \emph{Proceedings of the IEEE conference on computer vision and
  pattern recognition}, pp.\  1933--1941, 2016.

\bibitem[Gemmeke et~al.(2017)Gemmeke, Ellis, Freedman, Jansen, Lawrence, Moore,
  Plakal, and Ritter]{gemmeke2017audio}
Jort~F Gemmeke, Daniel~PW Ellis, Dylan Freedman, Aren Jansen, Wade Lawrence,
  R~Channing Moore, Manoj Plakal, and Marvin Ritter.
\newblock Audio set: An ontology and human-labeled dataset for audio events.
\newblock In \emph{2017 IEEE international conference on acoustics, speech and
  signal processing (ICASSP)}, pp.\  776--780. IEEE, 2017.

\bibitem[Gidaris et~al.(2018)Gidaris, Singh, and
  Komodakis]{gidaris2018unsupervised}
Spyros Gidaris, Praveer Singh, and Nikos Komodakis.
\newblock Unsupervised representation learning by predicting image rotations.
\newblock \emph{arXiv preprint arXiv:1803.07728}, 2018.

\bibitem[Girdhar et~al.(2023)Girdhar, El-Nouby, Liu, Singh, Alwala, Joulin, and
  Misra]{girdhar2023imagebind}
Rohit Girdhar, Alaaeldin El-Nouby, Zhuang Liu, Mannat Singh, Kalyan~Vasudev
  Alwala, Armand Joulin, and Ishan Misra.
\newblock Imagebind: One embedding space to bind them all.
\newblock In \emph{Proceedings of the IEEE/CVF Conference on Computer Vision
  and Pattern Recognition}, pp.\  15180--15190, 2023.

\bibitem[Gong et~al.(2021)Gong, Chung, and Glass]{gong2021ast}
Yuan Gong, Yu-An Chung, and James Glass.
\newblock Ast: Audio spectrogram transformer.
\newblock \emph{arXiv preprint arXiv:2104.01778}, 2021.

\bibitem[He et~al.(2016)He, Zhang, Ren, and Sun]{he2016deep}
Kaiming He, Xiangyu Zhang, Shaoqing Ren, and Jian Sun.
\newblock Deep residual learning for image recognition.
\newblock In \emph{Proceedings of the IEEE conference on computer vision and
  pattern recognition}, pp.\  770--778, 2016.

\bibitem[He et~al.(2022)He, Chen, Xie, Li, Doll{\'a}r, and
  Girshick]{he2022masked}
Kaiming He, Xinlei Chen, Saining Xie, Yanghao Li, Piotr Doll{\'a}r, and Ross
  Girshick.
\newblock Masked autoencoders are scalable vision learners.
\newblock In \emph{Proceedings of the IEEE/CVF conference on computer vision
  and pattern recognition}, pp.\  16000--16009, 2022.

\bibitem[He et~al.(2020)He, Liu, Gao, and Chen]{he2020deberta}
Pengcheng He, Xiaodong Liu, Jianfeng Gao, and Weizhu Chen.
\newblock Deberta: Decoding-enhanced bert with disentangled attention.
\newblock \emph{arXiv preprint arXiv:2006.03654}, 2020.

\bibitem[Houlsby et~al.(2019)Houlsby, Giurgiu, Jastrzebski, Morrone,
  De~Laroussilhe, Gesmundo, Attariyan, and Gelly]{houlsby2019parameter}
Neil Houlsby, Andrei Giurgiu, Stanislaw Jastrzebski, Bruna Morrone, Quentin
  De~Laroussilhe, Andrea Gesmundo, Mona Attariyan, and Sylvain Gelly.
\newblock Parameter-efficient transfer learning for nlp.
\newblock In \emph{International Conference on Machine Learning}, pp.\
  2790--2799. PMLR, 2019.

\bibitem[Hu et~al.(2021)Hu, Shen, Wallis, Allen-Zhu, Li, Wang, Wang, and
  Chen]{hu2021lora}
Edward~J Hu, Yelong Shen, Phillip Wallis, Zeyuan Allen-Zhu, Yuanzhi Li, Shean
  Wang, Lu~Wang, and Weizhu Chen.
\newblock Lora: Low-rank adaptation of large language models.
\newblock \emph{arXiv preprint arXiv:2106.09685}, 2021.

\bibitem[Huang et~al.(2022{\natexlab{a}})Huang, Xu, Li, Baevski, Auli, Galuba,
  Metze, and Feichtenhofer]{huang2022masked}
Po-Yao Huang, Hu~Xu, Juncheng Li, Alexei Baevski, Michael Auli, Wojciech
  Galuba, Florian Metze, and Christoph Feichtenhofer.
\newblock Masked autoencoders that listen.
\newblock \emph{arXiv preprint arXiv:2207.06405}, 2022{\natexlab{a}}.

\bibitem[Huang et~al.(2021)Huang, Du, Xue, Chen, Zhao, and
  Huang]{huang2021makes}
Yu~Huang, Chenzhuang Du, Zihui Xue, Xuanyao Chen, Hang Zhao, and Longbo Huang.
\newblock What makes multi-modal learning better than single (provably).
\newblock \emph{Advances in Neural Information Processing Systems},
  34:\penalty0 10944--10956, 2021.

\bibitem[Huang et~al.(2022{\natexlab{b}})Huang, Lin, Zhou, Yang, and
  Huang]{huang2022modality}
Yu~Huang, Junyang Lin, Chang Zhou, Hongxia Yang, and Longbo Huang.
\newblock Modality competition: What makes joint training of multi-modal
  network fail in deep learning?(provably).
\newblock In \emph{International Conference on Machine Learning}, pp.\
  9226--9259. PMLR, 2022{\natexlab{b}}.

\bibitem[Kiela et~al.(2019)Kiela, Bhooshan, Firooz, Perez, and
  Testuggine]{kiela2019supervised}
Douwe Kiela, Suvrat Bhooshan, Hamed Firooz, Ethan Perez, and Davide Testuggine.
\newblock Supervised multimodal bitransformers for classifying images and text.
\newblock \emph{arXiv preprint arXiv:1909.02950}, 2019.

\bibitem[Li et~al.(2023{\natexlab{a}})Li, Li, Savarese, and Hoi]{li2023blip}
Junnan Li, Dongxu Li, Silvio Savarese, and Steven Hoi.
\newblock Blip-2: Bootstrapping language-image pre-training with frozen image
  encoders and large language models.
\newblock \emph{arXiv preprint arXiv:2301.12597}, 2023{\natexlab{a}}.

\bibitem[Li et~al.(2023{\natexlab{b}})Li, Quan, Zhu, and Yang]{li2023efficient}
Yaowei Li, Ruijie Quan, Linchao Zhu, and Yi~Yang.
\newblock Efficient multimodal fusion via interactive prompting.
\newblock In \emph{Proceedings of the IEEE/CVF Conference on Computer Vision
  and Pattern Recognition}, pp.\  2604--2613, 2023{\natexlab{b}}.

\bibitem[Liu et~al.(2023{\natexlab{a}})Liu, Li, Wu, and Lee]{liu2023visual}
Haotian Liu, Chunyuan Li, Qingyang Wu, and Yong~Jae Lee.
\newblock Visual instruction tuning.
\newblock \emph{arXiv preprint arXiv:2304.08485}, 2023{\natexlab{a}}.

\bibitem[Liu et~al.(2021)Liu, Ji, Fu, Tam, Du, Yang, and Tang]{liu2021p}
Xiao Liu, Kaixuan Ji, Yicheng Fu, Weng~Lam Tam, Zhengxiao Du, Zhilin Yang, and
  Jie Tang.
\newblock P-tuning v2: Prompt tuning can be comparable to fine-tuning
  universally across scales and tasks.
\newblock \emph{arXiv preprint arXiv:2110.07602}, 2021.

\bibitem[Liu et~al.(2023{\natexlab{b}})Liu, Zheng, Du, Ding, Qian, Yang, and
  Tang]{liu2023gpt}
Xiao Liu, Yanan Zheng, Zhengxiao Du, Ming Ding, Yujie Qian, Zhilin Yang, and
  Jie Tang.
\newblock Gpt understands, too.
\newblock \emph{AI Open}, 2023{\natexlab{b}}.

\bibitem[Loshchilov \& Hutter(2017)Loshchilov and
  Hutter]{loshchilov2017decoupled}
Ilya Loshchilov and Frank Hutter.
\newblock Decoupled weight decay regularization.
\newblock \emph{arXiv preprint arXiv:1711.05101}, 2017.

\bibitem[Nagrani et~al.(2021)Nagrani, Yang, Arnab, Jansen, Schmid, and
  Sun]{nagrani2021attention}
Arsha Nagrani, Shan Yang, Anurag Arnab, Aren Jansen, Cordelia Schmid, and Chen
  Sun.
\newblock Attention bottlenecks for multimodal fusion.
\newblock \emph{Advances in Neural Information Processing Systems},
  34:\penalty0 14200--14213, 2021.

\bibitem[OpenAI(2023)]{openai2023gpt4}
OpenAI.
\newblock Gpt-4 technical report, 2023.

\bibitem[Oquab et~al.(2023)Oquab, Darcet, Moutakanni, Vo, Szafraniec, Khalidov,
  Fernandez, Haziza, Massa, El-Nouby, et~al.]{oquab2023dinov2}
Maxime Oquab, Timoth{\'e}e Darcet, Th{\'e}o Moutakanni, Huy Vo, Marc
  Szafraniec, Vasil Khalidov, Pierre Fernandez, Daniel Haziza, Francisco Massa,
  Alaaeldin El-Nouby, et~al.
\newblock Dinov2: Learning robust visual features without supervision.
\newblock \emph{arXiv preprint arXiv:2304.07193}, 2023.

\bibitem[Peng et~al.(2022)Peng, Wei, Deng, Wang, and Hu]{peng2022balanced}
Xiaokang Peng, Yake Wei, Andong Deng, Dong Wang, and Di~Hu.
\newblock Balanced multimodal learning via on-the-fly gradient modulation.
\newblock In \emph{Proceedings of the IEEE/CVF Conference on Computer Vision
  and Pattern Recognition}, pp.\  8238--8247, 2022.

\bibitem[Radford et~al.(2018)Radford, Narasimhan, Salimans, Sutskever,
  et~al.]{radford2018improving}
Alec Radford, Karthik Narasimhan, Tim Salimans, Ilya Sutskever, et~al.
\newblock Improving language understanding by generative pre-training.
\newblock 2018.

\bibitem[Radford et~al.(2021)Radford, Kim, Hallacy, Ramesh, Goh, Agarwal,
  Sastry, Askell, Mishkin, Clark, et~al.]{radford2021learning}
Alec Radford, Jong~Wook Kim, Chris Hallacy, Aditya Ramesh, Gabriel Goh,
  Sandhini Agarwal, Girish Sastry, Amanda Askell, Pamela Mishkin, Jack Clark,
  et~al.
\newblock Learning transferable visual models from natural language
  supervision.
\newblock In \emph{International conference on machine learning}, pp.\
  8748--8763. PMLR, 2021.

\bibitem[Saharia et~al.(2022)Saharia, Chan, Saxena, Li, Whang, Denton,
  Ghasemipour, Gontijo~Lopes, Karagol~Ayan, Salimans,
  et~al.]{saharia2022photorealistic}
Chitwan Saharia, William Chan, Saurabh Saxena, Lala Li, Jay Whang, Emily~L
  Denton, Kamyar Ghasemipour, Raphael Gontijo~Lopes, Burcu Karagol~Ayan, Tim
  Salimans, et~al.
\newblock Photorealistic text-to-image diffusion models with deep language
  understanding.
\newblock \emph{Advances in Neural Information Processing Systems},
  35:\penalty0 36479--36494, 2022.

\bibitem[Soomro et~al.(2012)Soomro, Zamir, and Shah]{soomro2012ucf101}
Khurram Soomro, Amir~Roshan Zamir, and Mubarak Shah.
\newblock Ucf101: A dataset of 101 human actions classes from videos in the
  wild.
\newblock \emph{arXiv preprint arXiv:1212.0402}, 2012.

\bibitem[Tian et~al.(2018)Tian, Shi, Li, Duan, and Xu]{tian2018audio}
Yapeng Tian, Jing Shi, Bochen Li, Zhiyao Duan, and Chenliang Xu.
\newblock Audio-visual event localization in unconstrained videos.
\newblock In \emph{Proceedings of the European conference on computer vision
  (ECCV)}, pp.\  247--263, 2018.

\bibitem[Touvron et~al.(2023)Touvron, Martin, Stone, Albert, Almahairi, Babaei,
  Bashlykov, Batra, Bhargava, Bhosale, et~al.]{touvron2023llama}
Hugo Touvron, Louis Martin, Kevin Stone, Peter Albert, Amjad Almahairi, Yasmine
  Babaei, Nikolay Bashlykov, Soumya Batra, Prajjwal Bhargava, Shruti Bhosale,
  et~al.
\newblock Llama 2: Open foundation and fine-tuned chat models.
\newblock \emph{arXiv preprint arXiv:2307.09288}, 2023.

\bibitem[Vaswani et~al.(2017)Vaswani, Shazeer, Parmar, Uszkoreit, Jones, Gomez,
  Kaiser, and Polosukhin]{vaswani2017attention}
Ashish Vaswani, Noam Shazeer, Niki Parmar, Jakob Uszkoreit, Llion Jones,
  Aidan~N Gomez, {\L}ukasz Kaiser, and Illia Polosukhin.
\newblock Attention is all you need.
\newblock \emph{Advances in neural information processing systems}, 30, 2017.

\bibitem[Wang et~al.(2020)Wang, Tran, and Feiszli]{wang2020makes}
Weiyao Wang, Du~Tran, and Matt Feiszli.
\newblock What makes training multi-modal classification networks hard?
\newblock In \emph{Proceedings of the IEEE/CVF conference on computer vision
  and pattern recognition}, pp.\  12695--12705, 2020.

\bibitem[Wang et~al.(2015)Wang, Kumar, Thome, Cord, and
  Precioso]{wang2015recipe}
Xin Wang, Devinder Kumar, Nicolas Thome, Matthieu Cord, and Frederic Precioso.
\newblock Recipe recognition with large multimodal food dataset.
\newblock In \emph{2015 IEEE International Conference on Multimedia \& Expo
  Workshops (ICMEW)}, pp.\  1--6. IEEE, 2015.

\bibitem[Wu et~al.(2022)Wu, Jastrzebski, Cho, and Geras]{wu2022characterizing}
Nan Wu, Stanislaw Jastrzebski, Kyunghyun Cho, and Krzysztof~J Geras.
\newblock Characterizing and overcoming the greedy nature of learning in
  multi-modal deep neural networks.
\newblock In \emph{International Conference on Machine Learning}, pp.\
  24043--24055. PMLR, 2022.

\bibitem[Xiao et~al.(2020)Xiao, Codevilla, Gurram, Urfalioglu, and
  L{\'o}pez]{xiao2020multimodal}
Yi~Xiao, Felipe Codevilla, Akhil Gurram, Onay Urfalioglu, and Antonio~M
  L{\'o}pez.
\newblock Multimodal end-to-end autonomous driving.
\newblock \emph{IEEE Transactions on Intelligent Transportation Systems},
  23\penalty0 (1):\penalty0 537--547, 2020.

\bibitem[Yang et~al.(2019)Yang, Dai, Yang, Carbonell, Salakhutdinov, and
  Le]{yang2019xlnet}
Zhilin Yang, Zihang Dai, Yiming Yang, Jaime Carbonell, Russ~R Salakhutdinov,
  and Quoc~V Le.
\newblock Xlnet: Generalized autoregressive pretraining for language
  understanding.
\newblock \emph{Advances in neural information processing systems}, 32, 2019.

\bibitem[Zhang et~al.(2023{\natexlab{a}})Zhang, Chen, Bukharin, He, Cheng,
  Chen, and Zhao]{zhang2023adaptive}
Qingru Zhang, Minshuo Chen, Alexander Bukharin, Pengcheng He, Yu~Cheng, Weizhu
  Chen, and Tuo Zhao.
\newblock Adaptive budget allocation for parameter-efficient fine-tuning.
\newblock \emph{arXiv preprint arXiv:2303.10512}, 2023{\natexlab{a}}.

\bibitem[Zhang et~al.(2023{\natexlab{b}})Zhang, Hu, Cui, Zhao, and
  Gao]{zhang2023universal}
Tong Zhang, Yingdong Hu, Hanchen Cui, Hang Zhao, and Yang Gao.
\newblock A universal semantic-geometric representation for robotic
  manipulation.
\newblock \emph{arXiv preprint arXiv:2306.10474}, 2023{\natexlab{b}}.

\end{thebibliography}
\bibliographystyle{iclr2024_conference}

\clearpage
\appendix
\section{Additional experimental settings}
\label{appendix_settings}
\subsection{Experimental settings on Audio Visual Datasets}
\paragraph{Settings of Uni-Modal Training.}
For uni-modal fine-tuning with large-scale pre-trained models, we add a linear layer for feature-to-label mapping and use a learning rate of $1e-5$. For training with ResNet18, the learning rate is set to $1e-4$.

\paragraph{Settings of Multi-Modal Training.}
% For the multi-modal joint training baseline, we set the learning rate to \(1e-5\).
% During the multi-modal adaptation phase of MMLoRA (with large-scale pre-trained models), we set the learning rate to \(1e-4\).
% When the backbones are resnet18,  we set the learning rate to \(1e-6\) of the multi-modal adaptation phase.
For multi-modal joint training, we use a learning rate of \(1e-5\). During MMLoRA's adaptation phase, the rate is \(1e-4\) with large uni-modal fine-tuned models, and \(1e-6\) with ResNet18 backbones.

\subsection{Experimental settings on Vision Language Datasets}
\textbf{Common Settings.} We reduce the learning rate when a metric has stopped improving. If the validation accuracy does not increase for two consecutive epochs, we reduce the learning rate to half of its original value.

\paragraph{Settings of Uni-Modal Training.}
For fine-tuning language models and ResNet, the learning rate is 5e-5. For ViT-L models, it's 1e-5. Noting that for vision-language datasets, ResNet152 is our baseline visual backbone.

\paragraph{Settings of Multi-Modal Training.}
For MMLoRA's multi-modal adaptation with large-scale uni-modal fine-tuned models, we use a learning rate of \(5e-4\) on UPMC Food101 and \(5e-6\) on MM-IMDB. With ResNet152 and BERT-base, it's \(1e-4\) on UPMC Food101 and \(1e-5\) on MM-IMDB.

\subsection{Experimental settings on Action Recognition Datasets~(UCF101)}
When training ViT-B on RGB data, we set the learning rate to \(1e-5\). For MMLoRA's multi-modal adaptation with ViT-B (RGB) and ResNet18 (optical flow), the rate is \(1e-4\). With ResNet18 for both RGB and optical flow, it's \(1e-5\). Other settings align with \citet{du2023uni}.

\section{Is bigger always better for models?}
\label{appendix_bigger}
When selecting more suitable pre-trained models for different datasets, we find that in some cases, bigger models do not necessarily yield better results. For instance, on the UPMC Food101 dataset, we select four different language models that vary both in pre-trained data and model size. However, their performance seems to be comparable as shown in Table~\ref{tab:food_text}. Thus, for the text encoder of this dataset, we used BERT-Base.
In CREMA-D, the smaller ViT-Base outperforms ViT-Large, as shown in Table~\ref{tab:CREMA-D_visual}. At the same time, ViT-B performs sufficiently well on other audio-visual datasets, even surpassing previous multi-modal methods. Therefore, we choose ViT-Base for the Audio-Visual datasets.

\begin{table}[h]
    \centering
    \caption{\textbf{Top-1 Test Accuracy of different language models fine-tuned on text modality of UPMC food101.}}
    \vspace{3pt}
    \begin{tabular}{c|c|c|c|c}
    \toprule
       \diagbox{Dataset}{Model}  & BERT-Base& BERT-Large & DeBERTa-Base & DeBERTa-Large\\
    \midrule
     UPMC food101    &  86.6 & 86.7 & 86.6 & 86.8\\
     
    \bottomrule
    \end{tabular}
    \label{tab:food_text}
\end{table}

\begin{table}[h]
    \centering
    \caption{\textbf{Top-1 Test Accuracy of different visual models fine-tuned with visual data of CREMA-D.}}
    \vspace{3pt}
    \begin{tabular}{c|c|c}
    \toprule
       \diagbox{Dataset}{Model}  & ViT-B & ViT-L\\
    \midrule
     CREMA-D    & 77.7 & 75.5\\
     
    \bottomrule
    \end{tabular}
    \label{tab:CREMA-D_visual}
\end{table}

\section{Additional ablation experiments}
\label{appendix_ablation}

\subsection{Implement MMLoRA directly on pre-trained models}
In this section, we conduct another ablation study on MMLoRA: directly reparametrizing the pre-trained models using LoRA and then proceeding with multi-modal joint training. Compared to the MMLoRA implementation in Sec~\ref{sec4}, this skips the uni-modal fine-tuning for the pre-trained models.
As shown in Table~\ref{tab:w_wo_UMFT}, MMLoRA with uni-modal fine-tuning performs much better than without it, which highlights the importance of uni-modal fine-tuning.

\begin{table}[t]
    \centering
    \caption{\textbf{Top-1 Test Accuracy of MMLoRA~(with UMFT and without UMFT) on AVE, Kinetics-Sound and CREMA-D}. We set rank of MMLoRA as 1 and UMFT represents Uni-Modal Fine-Tuning. Note that in this table, MMLoRA refers to reparametrizing both modality models using LoRA.}
    \vspace{3pt}
    \begin{tabular}{c|c|c|c}
    \toprule
       \diagbox{\textbf{Method}}{\textbf{Dataset}}  & \textbf{AVE}& \textbf{KS} & \textbf{CREMA-D}  \\
    \midrule
     MMLoRA~(w/o UMFT)    & 94.0 &89.5 & 83.7 \\
     MMLoRA~(w UMFT)    & \textbf{96.2}  & \textbf{91.4} &\textbf{88.3} \\
    \bottomrule
    \end{tabular}
    \label{tab:w_wo_UMFT}
\end{table}

\subsection{Reparameterize the linear layer with LoRA}

In this subsection, we conduct experiments to determine whether LoRA needs to be applied to the linear layer in MMLoRA. 
Specifically, we carry out the following experiments: \textit{1.} applying LoRA solely to the linear layer; \textit{2.} applying LoRA only to the encoder; and \textit{3.} applying LoRA to both the linear layer and the encoder. As shown in Table~\ref{tab:lora_which_part}, using LoRA to solely reparameterize the final Linear layers results in almost no gain on both datasets. Applying LoRA for reparameterization solely to the encoders seems to yield more stable results. Therefore, in the main experiments of our paper, we apply LoRA only to the encoders.

\begin{table}[t]
    \centering
    \caption{\textbf{Top-1 test accuracy of MMLoRA (when reparameterizing using LoRA on different parts) on AVE and Kinetics-Sound}. }
    \vspace{3pt}
    \begin{tabular}{c|c|c|c|c}
    \toprule
       \diagbox{\textbf{Dataset}}{\textbf{Method}}  & \textcolor{gray}{UME} & \textbf{Only Linear Layers}& \textbf{Only Encoders} & \textbf{Both}  \\
    \midrule
     AVE    & \textcolor{gray}{95.4} & 95.4 & 96.2 & 96.0 \\
     Kinetics-Sound & \textcolor{gray}{90.8}   & 90.8  & 91.4 & 91.4 \\
    \bottomrule
    \end{tabular}
    \label{tab:lora_which_part}
\end{table}
\end{document}